\documentclass{article}

\usepackage{PRIMEarxiv}

\usepackage[utf8]{inputenc} 
\usepackage[T1]{fontenc}    
\usepackage{hyperref}       
\usepackage{url}            
\usepackage{booktabs}       
\usepackage{amsfonts}       
\usepackage{nicefrac}       
\usepackage{microtype}      
\usepackage{lipsum}
\usepackage{amsmath} 
\usepackage{amssymb}
\usepackage{algorithm}
\usepackage{algpseudocode}
\usepackage{fancyhdr}       
\usepackage{graphicx}       
\usepackage{natbib}
\usepackage{float}
\usepackage{subfigure}
\usepackage{subcaption}
\usepackage{todonotes}

\graphicspath{{media/}}     

\title{A Financial Time Series Denoiser Based on Diffusion Model
}

\author{
  Zhuohan Wang \\
   Department of Informatics \\
  King's College London \\
  London, United Kingdom\\
  \texttt{zhuohan.wang@kcl.ac.uk}
  \And 
  Carmine Ventre \\
   Department of Informatics \\
  King's College London \\
  London, United Kingdom\\
  \texttt{carmine.ventre@kcl.ac.uk}
}

\begin{document}
\maketitle

\begin{abstract}
    Financial time series often exhibit low signal-to-noise ratio, posing significant challenges for accurate data interpretation and prediction and ultimately decision making. Generative models have gained attention as powerful tools for simulating and predicting intricate data patterns, with the diffusion model emerging as a particularly effective method. This paper introduces a novel approach utilizing the diffusion model as a denoiser for financial time series in order to improve data predictability and trading performance. By leveraging the forward and reverse processes of the conditional diffusion model to add and remove noise progressively, we reconstruct original data from noisy inputs. Our extensive experiments demonstrate that diffusion model-based denoised time series significantly enhance the performance on downstream future return classification tasks. Moreover, trading signals derived from the denoised data yield more profitable trades with fewer transactions, thereby minimizing transaction costs and increasing overall trading efficiency. Finally, we show that by using classifiers trained on denoised time series, we can recognize the noising state of the market and obtain excess return.
\end{abstract}


\section{Introduction}

In the realm of financial data analysis, the issue of low signal-to-noise ratio (SNR) is a critical problem that demands attention~\citep{israel2020can}. A low SNR indicates that the useful information (signal) is overwhelmed by random fluctuations or irrelevant data (noise). Noisy financial data can lead to misinterpretations and erroneous conclusions, potentially resulting in significant financial losses. On the other hand, noise in the data can degrade the performance of machine learning models, making their predictions less reliable and increasing the likelihood of poor decision outcomes.

Recently, generative models have become a pivotal tool in finance, leveraging their ability to simulate and predict complex data patterns. For example, generative models can be used for data generation~\citep{wiese2020quant, ni2020conditional, el2022styletime}, risk management~\citep{marti2020corrgan} and market simulation~\citep{byrd2019abides}. It makes one wonder whether generative models can be used for financial data denoising tasks. 

The diffusion model~\citep{ho2020denoising, song2019generative, song2020score}, as the most popular generative model at present, holds out hope in this context. It contains two data transformation processes, the forward process and reverse process. The forward process keeps adding noise to the original data until it reaches a completely noisy and unrecognizable state, often a prior Gaussian distribution. In the reverse process, the model learns to gradually remove the noise step by step, reconstructing the original data from the noisy version. In the computer vision research field, some recent papers use the diffusion model as a new adversarial purification method, which mitigates the vulnerability of images to adversarial attacks~\citep{nie2022diffusion, wang2022guided}. By leveraging a pre-trained diffusion model, they first add a small amount of noise to data in the forward process, then gradually purify adversarial images by the reverse process. Inspired by the success of the diffusion model in image datasets, we adopt the conditional diffusion model to denoise the financial time series and conduct extensive experiments on the denoised time series in practical trading settings.

We summarize our main contributions as follows.
$\mathbf{(1)}$ To mitigate the low SNR problem, we propose a conditional diffusion model with guidance to obtain denoised stock time series by the noising-denoising procedure in limited steps. The denoised time series shows better performance on the downstream return prediction task.
$\mathbf{(2)}$ We show that the denoised time series, obtained with our diffusion model, can generate more accurate trading signals on MACD and Bollinger strategies, with better return and fewer trading times.
$\mathbf{(3)}$ Using classifiers trained on denoised time series, we can obtain the denoised version of the market trend and distinguish whether the observed market trend is noisy. Based on this new information, we design new profitable trading strategies.

\section{Related Work}

We will elaborate on related research in three parts: the theory of diffusion models, the application of diffusion models on time series, and  denoising financial time series.

\noindent \textbf{Diffusion Model Theory.} There are two threads of diffusion models, which are eventually unified. The first thread of the diffusion model is inspired by thermodynamics~\citep{sohl2015deep}, laying the groundwork for diffusion models by proposing the idea of training generative models through gradual noise addition and removal. \citet{ho2020denoising} proposed the Denoising Diffusion Probabilistic Model (DDPM) in 2020, marking a significant milestone for diffusion models. Their method involves adding noise in a forward process and denoising in a reverse process to generate high-quality images. Subsequent work focuses on how to improve sampling quality and speed~\citep{nichol2021improved, song2020denoising}. 

The second thread of diffusion models originates from Score-based Generative Models (SGM)~\citep{song2019generative} by using Denoising Score Matching (DSM)~\citep{vincent2011connection}. It  involves training a model to predict the gradient of the data density (score) by minimizing the denoising score matching loss, which can be used for reconstructing clean data from noisy observations. \citet{song2019generative} showed that once the score function is learned, samples can be generated by iteratively refining noisy samples using Langevin dynamics. 

The above DDPM and SGM can be unified under the Stochastic Differential Equation (SDE) framework~\citep{song2020score}. This framework allows both training and sampling processes to be formulated using SDEs, providing a more elegant and theoretically grounded approach to diffusion modeling. Furthermore, \citet{karras2022elucidating} disentangle the complex design space of the diffusion model in the context of DSM, enabling significant improvements in sampling quality, training cost, and generation speed.

\noindent \textbf{Diffusion Model on Time Series.} The applications of diffusion models on time series can be divided into three categories~\citep{lin2023diffusion}, which are forecasting, imputation, and generation. TimeGrad~\citep{rasul2021autoregressive} is an auto-regressive time series forecasting model based on DDPM, while ScoreGrad~\citep{yan2021scoregrad} is developed on the SDE framework. For time series imputation task, CSDI~\citep{tashiro2021csdi} uses transformer~\citep{vaswani2017attention} as backbone model architecture, and SSSD~\citep{alcaraz2022diffusion} uses structured state space for sequence modeling method. For the time series generation task, DiffTime~\citep{coletta2024constrained} and its variants are proposed for constrained time series generation. $D^3$VAE~\citep{li2022generative} uses a coupled diffusion process for multivariate time series augmentation, then uses a bidirectional VAE with denoising score matching to clear the noise.

\noindent \textbf{Denoising Financial Time Series.} In \citep{song2021forecasting}, padding-based Fourier transform is proposed to eliminate the noise waveform in the frequency domain of financial time series data and recurrent neural networks are used to forecast future stock market index.  A hybrid model comprising of wavelet transform and optimized extreme learning machine is proposed in \citep{yu2020hybrid} to present a stable and precise prediction of financial time series. By using a Denoising AutoEncoder (DAE) architecture, \citet{ma2022denoised} try to denoise financial time series and demonstrate that denoised labels improve the performances of the downstream learning algorithm. 

\section{Score Based Generative Model: Background}\label{section:theory}
SGM is a self-supervised machine learning method to generate new data samples. In this section, we will briefly introduce the basics of SGM and two recent papers. Then, we show how these are unified under SDEs. 

Let $\{\mathbf{x}_i\}$ be a dataset from an unknown distribution $p_{data}(\mathbf{x})$. The \textit{score} of this unknown distribution is defined as $\nabla_{\mathbf{x}}\log p(\mathbf{x})$, which is a gradient of the log probability density function for data $\mathbf{x}$. If we know the score, we can start from a random data sample and move it towards the place with higher probability density via gradient ascent. We use a neural network $\mathbf{s}_{\theta}$ to approximate the score $\nabla_{\mathbf{x}}\log p(\mathbf{x})$. The training objective of $\mathbf{s}_{\theta}$ is to minimize the following objective:
\begin{equation}
\begin{split}
    \mathcal{L}(\theta) &= \frac{1}{2}\mathbb{E}_{p_{data}(x)}\big[||\mathbf{s}_{\theta}(\mathbf{x})- \nabla_{\mathbf{x}}\log p_{data}(\mathbf{x})||^2_2\big] \\
    &=\mathbb{E}_{p_{data}(x)}\big[tr(\nabla_{\mathbf{x}}\mathbf{s}_{\theta}(\mathbf{x}))+\frac{1}{2}||\mathbf{s}_{\theta}(\mathbf{x})||^2_2 \big] + const.
\end{split}
\label{eq:sgm-training}
\end{equation}
However, obtaining $tr(\nabla_{\mathbf{x}}\mathbf{s}_{\theta}(\mathbf{x}))$ in Equation \eqref{eq:sgm-training} needs a lot of compute. Thus, DSM~\citep{vincent2011connection} is used as a workaround to avoid the calculation of $tr(\nabla_{\mathbf{x}}\mathbf{s}_{\theta}(\mathbf{x}))$. It first perturbs data $\mathbf{x}$ by a predefined noise distribution $q_{\sigma}(\tilde{\mathbf{x}}|\mathbf{x})$ and then use DSM to estimate the score of noised data distribution $q_{\sigma}(\tilde{\mathbf{x}})\triangleq \int{q_{\sigma}(\tilde{\mathbf{x}}|\mathbf{x}) p_{data}(\mathbf{x})d\mathbf{x}}$. The training objective is to minimize the following:
\begin{equation}
    \frac{1}{2}\mathbb{E}_{q_{\sigma}(\tilde{\mathbf{x}}|\mathbf{x})p_{data}(\mathbf{x})}\big[||\mathbf{s}_{\theta}(\tilde{\mathbf{x}})- \nabla_{\tilde{\mathbf{x}}}\log q_{\sigma}(\tilde{\mathbf{x}}|\mathbf{x})||^2_2 \big]. 
\label{eq:dsm-training}
\end{equation}
It is worth mentioning that, under certain circumstances, the training objective of DSM is equivalent to that of DAE, which also follows a simple noising-denoising procedure~\citep{vincent2011connection}.

\subsection{Score Matching with Langevin Dynamics}

The process of Score Matching with Langevin Dynamics (SMLD) contains two parts. One is to use denoising score matching to estimate score of noised data distribution, the other is to use Langevin dynamics to sample from prior distribution iteratively. \citet{song2019generative} propose to perturb data with multiple-level noise and train a Noise Conditioned Score Network (NCSN) to estimate scores corresponding to all noise levels. The noising method is defined as $q_{\sigma}(\tilde{\mathbf{x}}|\mathbf{x})=\mathcal{N}(\tilde{\mathbf{x}}|\mathbf{x},\sigma^2\mathbf{I})$, where $\mathbf{I}$ is the identity matrix. The probability distribution of noised data is $q_{\sigma}(\tilde{\mathbf{x}})=\int p_{data}(\mathbf{x}) \mathcal{N}(\mathbf{\tilde{x}}|\mathbf{x}, \sigma^2 \mathbf{I})d\mathbf{x}$. They also define a noise sequence $\{\sigma_i\}^N_{i=1}$ satisfying $\sigma_{min} = \sigma_1<\sigma_2< \cdots<\sigma_N=\sigma_{max}$, where $\sigma_{min}$ is small enough such that $q_{\sigma_{min}}(\mathbf{\tilde{x}}) \approx p_{data}(\mathbf{x})$ and $\sigma_{max}$ is large enough such that $q_{\sigma_{max}}(\mathbf{\tilde{x}}) \approx \mathcal{N}(\mathbf{x}|0, \sigma^2_{max} \mathbf{I})$. To estimate the score of noised data distribution, 
a conditional score network $\mathbf{s}_{\mathbf{\theta}}(\mathbf{\tilde{x}}, \sigma)$ is trained to satisfy that $\forall\sigma \in \{\sigma_i\}^N_{i=1}:\mathbf{s}_{\mathbf{\theta}}(\mathbf{x}, \sigma) \approx \nabla_{\mathbf{x}}\log q_{\sigma}(\mathbf{x})$. The training objective of NCSN is a weighted sum of denoising score matching objectives, i.e., finding $\theta^*$ to minimise
\begin{equation}
    \sum_{i=1}^N \sigma_i^2 \mathbb{E}_{p_{data}(\mathbf{x})}\mathbb{E}_{q_{\sigma_i}(\tilde{\mathbf{x}}|\mathbf{x})}\big[||\mathbf{s}_{\theta^*}(\tilde{\mathbf{x}},\sigma_i)-\nabla_{\tilde{\mathbf{x}}}\log q_{\sigma_i}(\tilde{\mathbf{x}}|\mathbf{x})||^2_2 \big]. 
\label{eq:ncsn-training}
\end{equation}

SMLD adopts a Langevin MCMC sampling method to generate new data samples:
\begin{equation}
    \mathbf{x}_i^m=\mathbf{x}_i^{m-1}+\epsilon_i \mathbf{s}_{\theta^*}(\mathbf{x}_i^{m-1},\sigma_i)+\sqrt{2 \epsilon_i} \mathbf{z}_i^m, \ \ m=1, 2, \cdots, M
\label{eq:ncsn-sampling}
\end{equation}
where $\epsilon_i$ is the step size controled by $\{\sigma_i\}^N_{i=1}$ and $\mathbf{z}_i^m$ is standard Gaussian variable. The above sampling method needs to be iterated for $i=N, N-1, \cdots, 1$ with the starting point $x_N^0 \sim \mathcal{N}(\mathbf{x}|0, \sigma^2_{max} \mathbf{I})$ and $\mathbf{x}_i^0=\mathbf{x}_{i+1}^M$. As $M \rightarrow \infty$ and $\epsilon_i \rightarrow 0$, we will have $\mathbf{x}_1^M$ an exact sample from $q_{\sigma_{min}}(\mathbf{\tilde{x}}) \approx p_{data}(\mathbf{x})$.

\subsection{Denoising Diffusion Probabilistic Model}
Denoising Diffusion Probabilistic Model (DDPM)~\citep{sohl2015deep, ho2020denoising} can be seen as a hierachical markovian variational autoencoder~\citep{luo2022understanding}. Considering a noise sequence $0<\beta_1< \beta_2 <\cdots< \beta_N <1$ and forward noising process $p(\mathbf{x}_i|\mathbf{x}_{i-1})=\mathcal{N}(\mathbf{x}_i|\sqrt{1-\beta_i}\mathbf{x}_{i-1}, \beta_i\mathbf{I})$. Let $\alpha_i=\prod_{j=1}^i(1-\beta_j)$, then we have $p_{\alpha_i}(\mathbf{x}_i|\mathbf{x}_{0})=\mathcal{N}(\mathbf{x}_i|\sqrt{\alpha_i}\mathbf{x}_{0}, (1-\alpha_i)\mathbf{I})$. Similar to SMLD, the perturbed data distribution can be denoted as $p_{\alpha_i}(\tilde{\mathbf{x}})=\int p_{data}(\mathbf{x}) p_{\alpha_i}(\tilde{\mathbf{x}}|\mathbf{x})d\mathbf{x}$. The noise scale is predetermined to satisfy $p_{\alpha_N}(\tilde{\mathbf{x}}) \sim \mathcal{N}(\mathbf{0}, \mathbf{I})$. The reverse denoising process can be written as $p_{\theta}(\mathbf{x}_{i-1}|\mathbf{x}_i)=\mathcal{N}(\mathbf{x}_{i-1};\frac{1}{\sqrt{1-\beta_i}}(\mathbf{x}_i+\beta_i \mathbf{s}_{\theta}(\mathbf{x}_i,i)),\beta_i\mathbf{I})$. The training objective is a sum of weighted evidence lower bound (ELBO), i.e., finding $\theta^*$ that minimizes:
\begin{equation}
    \sum_{i=1}^N (1-\alpha_i) \mathbb{E}_{p_{data}(\mathbf{x})}\mathbb{E}_{p_{\alpha_i}(\tilde{\mathbf{x}}|\mathbf{x})}\big[||\mathbf{s}_{\theta^*}(\tilde{\mathbf{x}},i)-\nabla_{\tilde{\mathbf{x}}}\log p_{\alpha_i}(\tilde{\mathbf{x}}|\mathbf{x})||^2_2 \big].
\label{eq:ddpm-training}
\end{equation}
After solving Eq. (\ref{eq:ddpm-training}) we get score network $\mathbf{s}_{\mathbf{\theta}^*}(\mathbf{x}_i, i)$, the inference process follows:
\begin{equation}
    \mathbf{x}_{i-1}=\frac{1}{\sqrt{1-\beta_i}}(\mathbf{x}_i+\beta_i \mathbf{s}_{\mathbf{\theta}^*}(\mathbf{x}_i, i))+\sqrt{\beta_i} \mathbf{z}_i, \ \ \ i=N, N-1, \cdots, 1
\label{eq:ddpm-sampling}
\end{equation}
where $\mathbf{x}_N \sim \mathcal{N}(\mathbf{0}, \mathbf{I})$. This sampling method is called \textit{ancestral sampling}~\citep{song2020score}.

\subsection{Stochastic Differential Equation: A Unified Perspective}

\citet{song2020score} demonstrate that SMLD and DDPM can be unified under the perspective of SDEs. Let $\{\mathbf{x}(t) \}_{t=0}^T$ be a stochastic diffusion process indexed by a continuous time variable $t\in[0, T]$. $p_0$ is real data distribution and $p_T$ is 
tractable prior distribution such that $\mathbf{x}_0 \sim p_0$ and $\mathbf{x}_T \sim p_T$. They denote the probability density function of $\mathbf{x}(t)$ by $p_t(\mathbf{x})$ and the transition kernel from $\mathbf{x}(s)$ to $\mathbf{x}(t)$ by $p_{st}(\mathbf{x}(t)|\mathbf{x}(s))$ where $0\leq s<t \leq T$. Then, we can use a stochastic differential equation to represent such a forward diffusion process: 
\begin{equation}
    d\mathbf{x}=\mathbf{f}(\mathbf{x},t)\ dt + g(t) \ d \mathbf{w}
\label{eq:sde-forward-process}
\end{equation}
where $\mathbf{f}(\mathbf{x},t) dt$ is referred to as the \textit{drift} item, and $ g(t) d \mathbf{w}$ is referred to as the \textit{diffusion} item. $\mathbf{w}$ is a standard Wiener process and $d\mathbf{w} \sim \mathcal{N}(0, \mathbf{I})$. The synthetic data generation process is the reverse process of Eq. (\ref{eq:sde-forward-process}), which is also an SDE~\citep{anderson1982reverse}:
\begin{equation}
    d\mathbf{x}=[\mathbf{f}(\mathbf{x},t)-g^2(t) \nabla_{\mathbf{x}}\log p_t(\mathbf{x})]\ dt +  g(t)\ d \bar{\mathbf{w}}
\label{eq:sde-backward-process}
\end{equation}
where $\bar{\mathbf{w}}$ is a reverse-time Wiener process and  $\nabla_{\mathbf{x}}\log p_t(\mathbf{x})$ is the score of marginal distribution corresponding to each $t$. It starts from an initial data point $\mathbf{x}_T \sim p_T$ and gradually denoise it step by step following Eq. (\ref{eq:sde-backward-process}). Theoretically, if $dt$ is small enough, we can get $\mathbf{x}_0 \sim p_0$. To estimate $\nabla_{\mathbf{x}}\log p_t(\mathbf{x})$, the score network $\mathbf{s}_{\mathbf{\theta}}(\mathbf{x}, t)$ is trained with objective function: 
\begin{equation}
    \lambda(t) \mathbb{E}_t  \mathbb{E}_{\mathbf{x}_0} \mathbb{E}_{\mathbf{x}_t|\mathbf{x}_0}\big[||\mathbf{s}_{\theta}(\mathbf{x}(t),t)-\nabla_{\mathbf{x}(t)}\log p_{0t}(\mathbf{x}(t)|\mathbf{x}(0))||^2_2 \big]
\label{eq:sde-training}
\end{equation}
where $\lambda:[0:T] \rightarrow \mathbb{R}^+$ is a positive weight item and $t \sim \mathcal{U}[0,T]$. Eq. \eqref{eq:sde-training} is a continuous generalization of Eq. \eqref{eq:ncsn-training} and Eq. \eqref{eq:ddpm-training}. In \citep{song2020score}, the forward process of SMLD and DDPM can be seen as a discrete form of continuous-time SDEs. Typically, the discrete forward process of SMLD is:
\begin{equation}
    \mathbf{x}_i=\mathbf{x}_{i-1} + \sqrt{\sigma_i^2-\sigma_{i-1}^2} \mathbf{z}_{i-1}, \ \ i = 1, \cdots , N.
\label{smld-discrete-forward-process}
\end{equation}
When $N \rightarrow \infty$, the discrete form $\{\mathbf{x}_i\}_{i=1}^N$ becomes continuous form $\{\mathbf{x}(t)\}_{t=0}^1$; and the continuous forward process can be written as Eq. (\ref{smld-continuous-forward-process}):
\begin{equation}
    d\mathbf{x}=\sqrt{\frac{d[\sigma(t)^2]}{dt}} \ d\mathbf{w}
\label{smld-continuous-forward-process}
\end{equation}
where $\mathbf{f}(\mathbf{x},t)=0$ and $g(t) = \sqrt{\frac{d[\sigma(t)^2]}{dt}}$. This is called Variance Exploding (VE) SDE.

As for DDPM, the discrete forward process is:
\begin{equation}
    \mathbf{x}_i=\sqrt{1-\beta_i}\mathbf{x}_{i-1}+\sqrt{\beta_i}\mathbf{z}_{i-1}, \ \ i = 1, \cdots , N 
\label{ddpm-discrete-forward-process}
\end{equation}
When $N \rightarrow \infty$, the continuous form of DDPM forward process can be written as Eq. (\ref{ddpm-continuous-forward-process}):
\begin{equation}
    d\mathbf{x}=-\frac{\beta(t)}{2} \mathbf{x}\  dt + \sqrt{\beta(t)} \ d\mathbf{w}
\label{ddpm-continuous-forward-process}
\end{equation}
where $\mathbf{f}(\mathbf{x},t)=-\frac{\beta(t)}{2} \mathbf{x}$ and $g(t) = \sqrt{\beta(t)}$. This is called Variance Preserving (VP) SDE. By substituting $\mathbf{f}(\mathbf{x},t)$ and $g(t)$ in Eq. (\ref{eq:sde-backward-process}) with $\mathbf{f}(\mathbf{x},t)$ and $g(t)$ in Eq. (\ref{smld-continuous-forward-process}) (Eq. (\ref{ddpm-continuous-forward-process}), respectively), we can get the corresponding backward process of SDE form for SMLD (DDPM, respectively) .

\section{Methodology}\label{section:methodology}

In this section, we will show how to train the conditional diffusion model by stock time series datasets under the SDE framework shown in Section \ref{section:theory}, and how we can get denoised stock time series via trained diffusion model. 

\subsection{Problem Formulation}
Let $\mathbf{x}$ and $\mathbf{c}$ be two time series of length $L$, where $\mathbf{x} \in \mathbb{R}^L$ is the input to our diffusion model and $\mathbf{c} \in \mathbb{R}^L$ is the model's condition. The size of denoised time series $\hat{\mathbf{x}}$ is also $\mathbb{R}^L$. 

A neural network $\mathbf{s}_{\theta}(\mathbf{x}, t, \mathbf{c})$ is used to estimate the score of noised data distribution. Here we use the network architecture from CSDI \citep{tashiro2021csdi}, which is a conditional transformer-based neural network, to train $\mathbf{s}_{\theta}(\mathbf{x}, t, \mathbf{c})$. It takes time series $\mathbf{x}$ as input, time series $\mathbf{c}$ and sinusoidal-embedded time $t$ as the condition. 

\subsection{Training and Inference}

In the training stage, we follow the training method in \citep{song2020score} where a random level of noise $t \sim \mathcal{U}(0, T)$ is added to the input data $\mathbf{x}_0$ via forward diffusion process shown in Eq. (\ref{eq:sde-forward-process}), then the network parameter $\mathbf{\theta}$ is updated by Eq. (\ref{eq:sde-training}) until convergence. It should be noted that during training the noise level $t$ is uniformly sampled from $[0, T]$, where $T=1$ is the maximum noise level. To train conditional diffusion model, we adopt classifier-free guidance method which is explained below. Next, we will show how to use trained conditional diffusion model to denoise a given time series. 

\begin{figure}[h]  
    \centering  
    \includegraphics[width=0.6\linewidth]{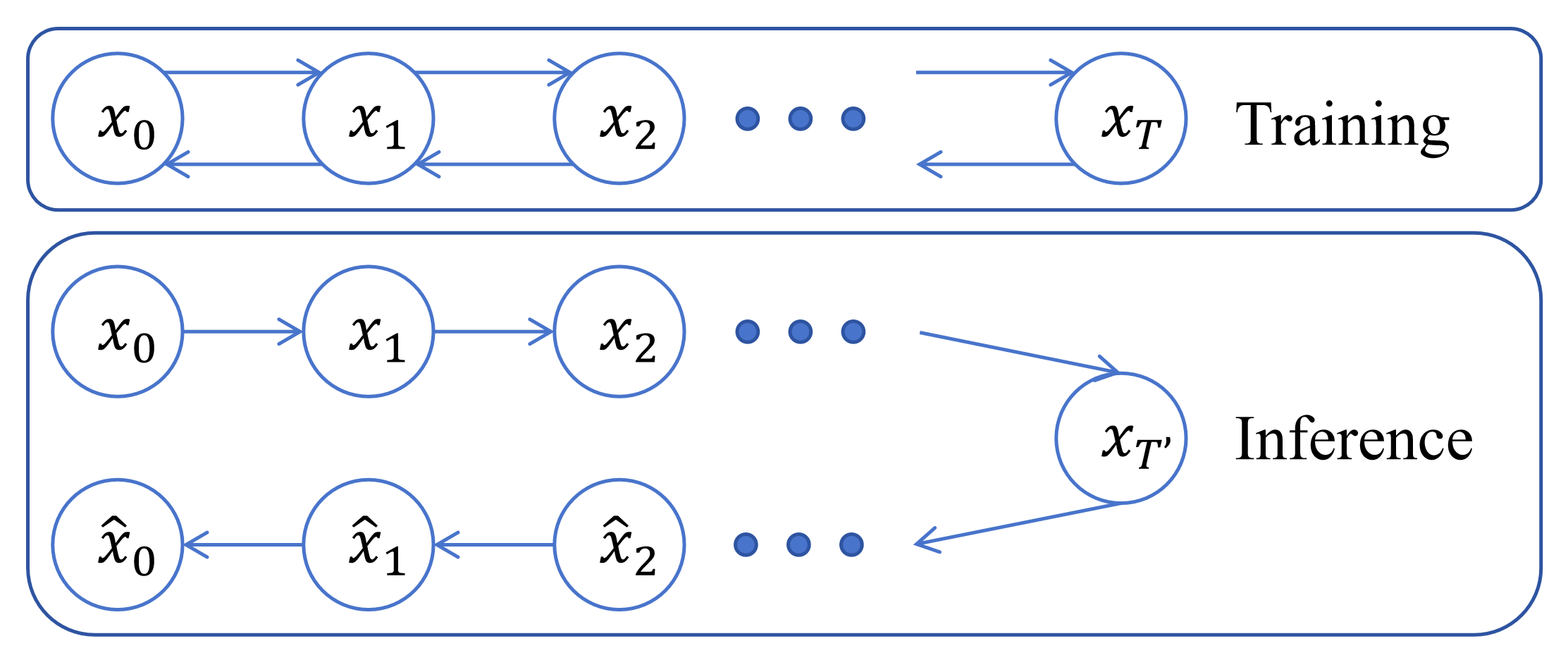}  
    \caption{Illustration of Training and Inference.}
    \label{fig:noising-denoising}  
\end{figure}

\noindent \textbf{Noising-Denoising Procedure.} To denoise time series, we adopt the noising-denoising procedure in the inference stage. Since we aim to denoise the input data instead of generating new data, we should ensure that the denoised data still contains similar moving trends as input data. Therefore, our initial data should be $\mathbf{x}_0$ instead of $\mathbf{x}_T \in \mathcal{N}(0, \mathbf{I})$ in the inference stage. By using a certain noise level $T' < T$, we can obtain a noised data $\mathbf{x}_{T'}$. Notice that $T'$ should be large enough so that the noise structure of $\mathbf{x}_0$ is removed but not too large to destroy the intrinsic trend. Although $T'$ is often set as a hyperparameter in previous work~\citep{nie2022diffusion, gao2024diffsformer} because it cannot be theoretically determined, we add the extra condition $\mathbf{c}=\mathbf{x}_0$ in the training and inference stage so that the choice of $T'$ does not have much influence on denoising performance. By letting $\mathbf{c}=\mathbf{x}_0$, the noised data $\mathbf{x}_t$ will always move towards the original $\mathbf{x}_0$ no matter what noise level $T'$ we choose. The illustration of training and inference is shown in Figure \ref{fig:noising-denoising}. 

\noindent \textbf{Classifier-free Guidance.} How do we inject condition $\mathbf{c}$ into the training and inference process? Here we take \textit{classifier-free guidance} approach~\citep{ho2022classifier}. It is developed on \textit{classifier guidance}~\citep{dhariwal2021diffusion}, where the conditional score $\nabla_x \log p(x|c)$ is calculated by:
\begin{equation}
    \nabla_x \log p(x|c)=\nabla_x \log p(x)+\omega\nabla_x \log p(c|x)
\label{eq:classifier-guidance}
\end{equation}
where $\nabla_x \log p(x)$ is unconditional score and $\nabla_x \log p(c|x)$ is the gradient of classifier; $\omega$ is a hyperparameter controlling the guidance strength. Doing so can decrease the diversity of generated data while increasing sampling quality. To avoid training an extra classifier, classifier-free guidance combines the conditional model and unconditional model:
\begin{equation}
    \nabla_x \log \tilde{p}(x|c)=\omega\nabla_x \log p(x|c)+(1-\omega)\nabla_x \log p(x)
\label{eq:classifier-free-guidance}
\end{equation}
where $\nabla_x \log p(x|c)$ represents the sampling direction of conditional model and $\nabla_x \log p(x)$ represents the sampling direction of unconditional model. Eq. (\ref{eq:classifier-free-guidance}) degrades to the score of the unconditional model when $\omega=0$, or becomes the score of the conditional model when $\omega=1$. By tuning $\omega$, we can make our model more flexible.

\noindent \textbf{Guidance by two loss function.} To increase the predictability of financial time series, we consider two kinds of loss functions as extra guidance during the inference process. The first loss function is \textit{Total Variation (TV) Loss}, which is defined as:
\begin{equation}
    \mathcal{L}_{\text{TV}}(\mathbf{x}) = \sum_{i=1}^{L-1}|\mathbf{x}_{[i+1]}-\mathbf{x}_{[i]}|
\label{eq:tv-loss}
\end{equation}
where $\mathbf{x}_{[i]}$ represents the i-th value of $\mathbf{x}$. TV Loss is a widely used regularization term for image denoising, deblurring, and super-resolution~\citep{allard2008total}. The main goal of TV Loss is to promote smoothness in the reconstructed data by penalizing rapid changes between neighboring values.  

The second loss function is \textit{Fourier Loss}, defined as:
\begin{equation}
    \mathcal{L}_{\text{F}}(\mathbf{x}_t, \mathbf{x}) = ||\mathcal{FFT}(\mathbf{x}_t)-Filter(\mathcal{FFT}(\mathbf{x}), f)||_2^2.
\label{eq:fft-loss}
\end{equation}
Fourier Loss is a technique used to ensure that the frequency domain characteristics of a given signal match a desired target~\citep{gopinathan2015wavelet}. Here $\mathbf{x}_t$ is noised data and $\mathbf{x}$ is original data. $\mathcal{FFT}(\cdot)$ represents Fast Fourier Transform (FFT) which converts data from time domain to frequency domain. We assume that the noise mostly exists in those frequencies with low amplitude, so we use $Filter(\cdot)$ to set frequencies whose amplitude is below a certain threshold $f$ to zero. Experimentally, we set the threshold $f$ to be $0.1$. 

\noindent \textbf{Predictor-Corrector Sampler} To better improve our solution to reverse SDE, we can use Eq. (\ref{eq:sde-backward-process}) as the initial solution and employ score-based MCMC approaches to correct the solution at every step, which is called Predictor-Corrector samplers proposed in paper~\citep{song2020score}. For VE-SDE, the Predictor is a discrete reverse process step: $\mathbf{x}_{i-1}=\mathbf{x}_i + (\sigma_i^2-\sigma_{i-1}^2)\nabla_{\mathbf{x}}\log p_i(\mathbf{x}_i) + \sqrt{(\sigma_i^2-\sigma_{i-1}^2)}\mathbf{z}_i$, while for VP-SDE, the Predictor is a different discrete reverse process step: $\mathbf{x}_{i-1}=\frac{1}{\sqrt{1-\beta_i}} [\mathbf{x}_i+\frac{\beta_i}{2}\nabla_{\mathbf{x}}\log p_i(\mathbf{x}_i)]+\sqrt{\beta_i} \mathbf{z}_i$, where $\mathbf{z}_i \sim \mathcal{N}(0, \mathbf{I})$. For both VE-SDE and VP-SDE, we choose annealed Langevin MCMC\citep{song2019generative} as our Corrector.

\noindent \textbf{Reduce Sampling Randomness.} Due to the randomness of the reverse denoising process, the denoised time series sometimes might still deviate from the original time series a lot, leading to information loss. To overcome the randomness, we use multiple independent reverse processes to denoise data by $s$ random seeds, then average all the denoised data samples to get the final denoised time series $\mathbf{\hat{x}}$. 

Compared to other self-supervised learning-based denoisers such as DAE~\citep{ma2022denoised}, our diffusion model-based denoising framework is more flexible due to multiple noising-denoising steps and auxiliary guidance during inference. The complete training and inference algorithm are shown in Algorithms \ref{algo:training} and \ref{algo:sampling}.

\setcounter{algorithm}{0}
\begin{algorithm}
\caption{Training Algorithm}\label{algo:training}
\begin{algorithmic}[1]
\Require data $\mathbf{x}_0$, condition $\mathbf{c} = \varnothing$ if training unconditional model else $\mathbf{c} = \mathbf{x}_0$
\State Initialize model parameters \(\theta\)

\For{each training step}
    \State Sample time \(t \sim \mathcal{U}(0, T)\)
    \State Get perturbed data $\mathbf{x}_t$ by $ p_{0t}(\mathbf{x}_t|\mathbf{x}_0)$ 
    \State Update parameters $\theta$ by Eq. (\ref{eq:sde-training})
\EndFor
\Ensure Network $\mathbf{s}_{\theta}(\mathbf{x}, t, \mathbf{c})$

\end{algorithmic}
\end{algorithm}

\begin{algorithm}[tbh]
\caption{Inference Algorithm}\label{algo:sampling}
\begin{algorithmic}[1]
\Require data $\mathbf{x}_0$, noising step $T'$, corrector step $M$, condition $\mathbf{c}$, empty list $\mathbf{x}_{list}=[]$
\State Divide $[0, T]$ into $N$ sections, so that
$t^{i=1, 2, \cdots, N}=\frac{i \cdot T}{N}$
\State Choose $T' \in [0, T]$ and calculate $K=\frac{N \cdot T'}{T}$
\For{random seed = $1$ to $s$}
\State Get perturbed data $\mathbf{x}_K$ by $ p_{0T'}(\mathbf{x}_{T'}|\mathbf{x}_0)$ 
\For{i = $K-1$ to $0$}
    \State $\mathbf{x}_i \leftarrow$ Predictor$(\mathbf{x}^{i+1}, \mathbf{t}^{i+1}, \mathbf{c})$
    \For{j = $1$ to $M$}
    \State $\mathbf{x}_i \leftarrow$ Corrector$(\mathbf{x}^{i+1}, \mathbf{t}^{i+1}, \mathbf{c})$
    \EndFor    
    \State $\mathbf{x}_i \leftarrow \mathbf{x}_i-\eta_{TV} \nabla_{\mathbf{x}_i} \mathcal{L}_{TV}(\mathbf{x}_i)$
    \State $\mathbf{x}_i \leftarrow \mathbf{x}_i-\eta_{F} \nabla_{\mathbf{x}_i} \mathcal{L}_{F}(\mathbf{x}_i, \mathbf{c})$
\EndFor
\State Add $\mathbf{x}_0$ to $\mathbf{x}_{list}$
\EndFor
\State $\mathbf{\hat{x}} \leftarrow$ Mean($\mathbf{x}_{list}$)
\Ensure denoised data $\hat{\mathbf{x}}$

\end{algorithmic}
\end{algorithm}

\section{Experiments}
In this section, we will first introduce the datasets we use in the following experiments. Then, we show the diffusion model based on the denoised stock price time series in Section \ref{subsection:diffusion-model-based-denoised-time-series}. Next, we try to answer the following three questions in our experiments: $\mathbf{(1)}$ Does our diffusion model-based denoiser increase the signal-to-noise ratio of stock time series? $\mathbf{(2)}$ Can our diffusion model-based denoiser be used for trading and obtain greater profits? $\mathbf{(3)}$ What kind of trading strategies can we develop if we are able to distinguish whether the current market trend is noisy or not?

\subsection{Datasets}\label{subsection:datasets-and-hyperparameters}
We use three stock datasets of 1day (2014.01.01-2023.12.31), 1hour (2023.06.07-2024.04.02) and 5min (2024.02.12-2024.04.02) frequency from Stooq website~\footnote{https://stooq.com/db/h/}. We only keep the US S\&P 500 stocks as our stock pool. For each stock in the stock pool, we use rolling window of size 60 and stride 20 to get stock closing price time series, so that each data sample in the dataset is of length 60. Each dataset is divided into training and testing periods in the proportion of $4:1$. The number of time series contained in each training dataset is 47,807, 25,985, and 51,794, respectively.

\subsection{Diffusion Model Based Denoised Time Series}\label{subsection:diffusion-model-based-denoised-time-series}
\begin{figure*}[tb]  
    \centering  
    \includegraphics[width=\textwidth]{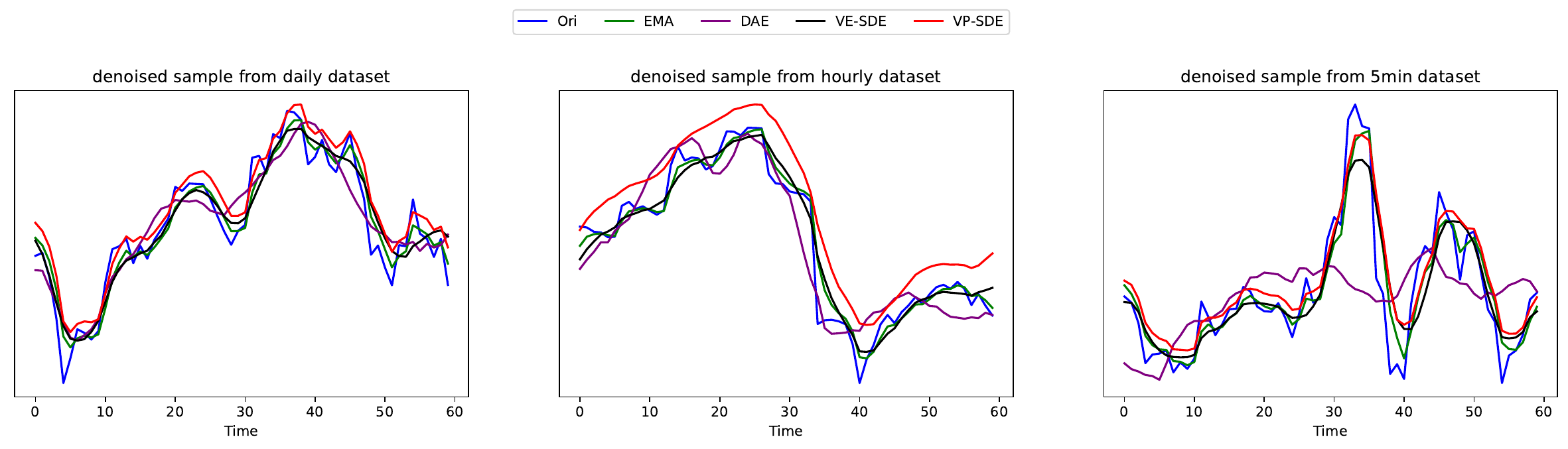}  
    \caption{Denoised Time Series Samples from 1day (left), 1hour (middle) and 5min (right) Dataset.}
    \label{fig:denoised-samples}  
\end{figure*}

We use Ori to represent the original closing price time series and EMA for the Exponential Moving Average of Ori, where the decay factor is $0.5$. Let the EMA be the diffusion model input $\mathbf{x}_0$. The classifier-free guidance scale $\omega$ is set to be $1$. By exploiting the training algorithm and inference algorithm, we can obtain VE-SDE and VP-SDE-based denoised time series. We also consider DAE as the benchmark denoising model~\citep{ma2022denoised}. The denoised time series samples of three datasets are shown in Figure \ref{fig:denoised-samples}. 

To better observe the characteristics of these time series, we use the number of directional change events~\citep{guillaume1997bird} as a tool. Directional change events focus on discrete points where the price movement is substantial enough to indicate a change in the trend's direction. When the price moves by this threshold in one direction, it triggers a directional change event. We consider this to be a measure of the ``market clock'' and a parameter to assess the quality of synthetic time series. In Table~\ref{tab:dc-events}, we show the average number of directional change events of different time series under various thresholds. We use thresholds taken from $[0.01, 0.02, 0.03]$ for 1day and 1hour datasets, and $[0.001, 0.002, 0.003]$ for 5min dataset. We observe that (1) DAE has the least DC events in most cases; (2) In the 1hour dataset, VE/VP-SDE has more DC events than EMA.

\begin{table*}\centering
  
  \scalebox{1}{
    \begin{tabular}{l|ccc|ccc|ccc|}
    \cline{2-10}
                & \multicolumn{3}{c|}{1day}                                      & \multicolumn{3}{c|}{1hour}                                   & \multicolumn{3}{c|}{5min}                                       \\ \cline{2-10} 
                                 & \multicolumn{1}{c|}{0.01}  & \multicolumn{1}{c|}{0.02}  & 0.03 & \multicolumn{1}{c|}{0.01} & \multicolumn{1}{c|}{0.02} & 0.03 & \multicolumn{1}{c|}{0.001} & \multicolumn{1}{c|}{0.002} & 0.003 \\ \hline
    \multicolumn{1}{|l|}{Ori}    & \multicolumn{1}{c|}{18.15} & \multicolumn{1}{c|}{11.80} & 8.32 & \multicolumn{1}{c|}{6.16} & \multicolumn{1}{c|}{3.08} & 2.07 & \multicolumn{1}{c|}{13.45} & \multicolumn{1}{c|}{7.47}  & 4.95  \\ \hline
    \multicolumn{1}{|l|}{EMA}    & \multicolumn{1}{c|}{8.96}  & \multicolumn{1}{c|}{6.04}  & 4.7  & \multicolumn{1}{c|}{3.76} & \multicolumn{1}{c|}{2.29} & 1.66 & \multicolumn{1}{c|}{6.63}  & \multicolumn{1}{c|}{4.13}  & 3.03  \\ \hline
    \multicolumn{1}{|l|}{DAE}    & \multicolumn{1}{c|}{5.72}  & \multicolumn{1}{c|}{4.45}  & 3.72 & \multicolumn{1}{c|}{2.52} & \multicolumn{1}{c|}{1.82} & 1.49 & \multicolumn{1}{c|}{7.24}  & \multicolumn{1}{c|}{3.92}  & 2.62  \\ \hline
    \multicolumn{1}{|l|}{VE-SDE} & \multicolumn{1}{c|}{5.98}  & \multicolumn{1}{c|}{4.89}  & 4.14 & \multicolumn{1}{c|}{4.04} & \multicolumn{1}{c|}{2.51} & 1.81 & \multicolumn{1}{c|}{4.46}  & \multicolumn{1}{c|}{3.36}  & 2.74  \\ \hline
    \multicolumn{1}{|l|}{VP-SDE} & \multicolumn{1}{c|}{6.50}  & \multicolumn{1}{c|}{5.12}  & 4.26 & \multicolumn{1}{c|}{3.81} & \multicolumn{1}{c|}{2.43} & 1.76 & \multicolumn{1}{c|}{5.00}  & \multicolumn{1}{c|}{3.59}  & 2.82  \\ \hline
    \end{tabular}}
\caption{Directional Change(DC) Events.}
\label{tab:dc-events}
\end{table*}

\subsection{Downstream Classification Task} \label{subsection:downstream-classification-task}
To evaluate if the signal-to-noise ratio of data is increased after the denoising process, we measure the predictability of denoised time series in downstream tasks. We set a supervised binary classification task to predict the future log return. It should be pointed out that the labels used here are obtained by corresponding time series, e.g., the label of Ori data is calculated by Ori data, and the label of denoised data is calculated by denoised data. We use an ensemble-tree-based model as our classifier~\citep{ke2017lightgbm} due to its robustness and quickness. The input of the classification model is a 1D time series $\mathbf{x}_{[t-59, t]}$, which is of length 60. The output is 0 (negative and zero log return) or 1 (positive log return). We perform experiments on three prediction lengths of $[t, t+1]$, $[t,t+5]$ and $[t,t+10]$, representing 1, 5, and 10 timesteps prediction, respectively. 


To assess the prediction performance, we use F1 Score and Matthews Correlation Coefficient (MCC) as evaluation metrics. The metric calculation methods are shown in Eq. (\ref{eq:f1score}) and Eq. (\ref{eq:mcc}):
{\allowdisplaybreaks
\begin{align}
    \text{F1 Score} & = 2 \cdot \frac{\left(\frac{TP}{TP + FP}\right) \cdot \left(\frac{TP}{TP + FN}\right)}{\left(\frac{TP}{TP + FP}\right) + \left(\frac{TP}{TP + FN}\right)}
\label{eq:f1score}\\
    \text{MCC} & = \frac{TP \cdot TN - FP \cdot FN}{\sqrt{(TP + FP)(TP + FN)(TN + FP)(TN + FN)}}.
\label{eq:mcc}
\end{align}
}
To measure the effect of noising step $T'$ on model performance, we choose different noising steps $T' \in \frac{[100, 200, \cdots, 900]}{1000} \times T$ for VE-SDE and VP-SDE, which means we can get $9$ prediction results for each diffusion model. In the experiments, the noising step does not have an obvious effect on model performance, which means our conditional diffusion model does not need a specially designed noising step. We use 5 different random seeds to get an average performance on each task. The best performance of each method on F1 Score and MCC on 3 datasets are shown in Table \ref{tab:1day-f1-mcc}, Table \ref{tab:1hour-f1-mcc} and Table \ref{tab:5min-f1-mcc}, respectively. We can find that our proposed VE-SDE and VP-SDE-based denoising diffusion model can beat DAE and EMA-based denoising methods in most prediction cases.

\begin{table*}\centering
\scalebox{1}{
\begin{tabular}{l|cccccc|}
\cline{2-7}
                             & \multicolumn{6}{c|}{1day}                                                                                                                                                                                    \\ \cline{2-7} 
                             & \multicolumn{2}{c|}{1 step}                                               & \multicolumn{2}{c|}{5 steps}                                              & \multicolumn{2}{c|}{10 steps}                        \\ \cline{2-7} 
                             & \multicolumn{1}{c|}{F1}             & \multicolumn{1}{c|}{MCC}            & \multicolumn{1}{c|}{F1}             & \multicolumn{1}{c|}{MCC}            & \multicolumn{1}{c|}{F1}             & MCC            \\ \hline
\multicolumn{1}{|l|}{Ori}    & \multicolumn{1}{c|}{0.604}          & \multicolumn{1}{c|}{0.014}          & \multicolumn{1}{c|}{0.641}          & \multicolumn{1}{c|}{0.036}          & \multicolumn{1}{c|}{0.634}          & 0.017          \\ \hline
\multicolumn{1}{|l|}{EMA}    & \multicolumn{1}{c|}{0.656}          & \multicolumn{1}{c|}{0.120}          & \multicolumn{1}{c|}{0.660}          & \multicolumn{1}{c|}{0.070}          & \multicolumn{1}{c|}{0.652}          & 0.025          \\ \hline
\multicolumn{1}{|l|}{DAE}    & \multicolumn{1}{c|}{0.676}          & \multicolumn{1}{c|}{0.183}          & \multicolumn{1}{c|}{0.685}          & \multicolumn{1}{c|}{0.129}          & \multicolumn{1}{c|}{0.662}          & 0.064          \\ \hline
\multicolumn{1}{|l|}{VE-SDE} & \multicolumn{1}{c|}{\textbf{0.720}} & \multicolumn{1}{c|}{\textbf{0.332}} & \multicolumn{1}{c|}{\textbf{0.731}} & \multicolumn{1}{c|}{\textbf{0.445}} & \multicolumn{1}{c|}{\textbf{0.668}} & \textbf{0.281} \\ \hline
\multicolumn{1}{|l|}{VP-SDE} & \multicolumn{1}{c|}{0.719}          & \multicolumn{1}{c|}{0.323}          & \multicolumn{1}{c|}{0.706}          & \multicolumn{1}{c|}{0.393}          & \multicolumn{1}{c|}{0.648}          & 0.230          \\ \hline
\end{tabular}}
\caption{Performance of 1day Dataset on F1-Score and MCC Metrics.}
\label{tab:1day-f1-mcc}
\end{table*}

\begin{table*}\centering
\scalebox{1}{
\begin{tabular}{l|cccccc|}
\cline{2-7}
                             & \multicolumn{6}{c|}{1hour}                                                                                                                                                                                   \\ \cline{2-7} 
                             & \multicolumn{2}{c|}{1 step}                                               & \multicolumn{2}{c|}{5 steps}                                              & \multicolumn{2}{c|}{10 steps}                        \\ \cline{2-7} 
                             & \multicolumn{1}{c|}{F1}             & \multicolumn{1}{c|}{MCC}            & \multicolumn{1}{c|}{F1}             & \multicolumn{1}{c|}{MCC}            & \multicolumn{1}{c|}{F1}             & MCC            \\ \hline
\multicolumn{1}{|l|}{Ori}    & \multicolumn{1}{c|}{0.452}          & \multicolumn{1}{c|}{-0.003}         & \multicolumn{1}{c|}{0.493}          & \multicolumn{1}{c|}{-0.001}         & \multicolumn{1}{c|}{0.521}          & -0.012         \\ \hline
\multicolumn{1}{|l|}{EMA}    & \multicolumn{1}{c|}{0.558}          & \multicolumn{1}{c|}{0.085}          & \multicolumn{1}{c|}{0.526}          & \multicolumn{1}{c|}{0.002}          & \multicolumn{1}{c|}{0.563}          & -0.008         \\ \hline
\multicolumn{1}{|l|}{DAE}    & \multicolumn{1}{c|}{0.563}          & \multicolumn{1}{c|}{0.065}          & \multicolumn{1}{c|}{0.499}          & \multicolumn{1}{c|}{0.030}          & \multicolumn{1}{c|}{0.604}          & 0.031          \\ \hline
\multicolumn{1}{|l|}{VE-SDE} & \multicolumn{1}{c|}{0.774}          & \multicolumn{1}{c|}{0.291}          & \multicolumn{1}{c|}{0.681}          & \multicolumn{1}{c|}{0.192}          & \multicolumn{1}{c|}{0.691}          & 0.131          \\ \hline
\multicolumn{1}{|l|}{VP-SDE} & \multicolumn{1}{c|}{\textbf{0.806}} & \multicolumn{1}{c|}{\textbf{0.329}} & \multicolumn{1}{c|}{\textbf{0.762}} & \multicolumn{1}{c|}{\textbf{0.350}} & \multicolumn{1}{c|}{\textbf{0.740}} & \textbf{0.238} \\ \hline
\end{tabular}}
\caption{Performance of 1hour Dataset on F1-Score and MCC Metrics.}
\label{tab:1hour-f1-mcc}
\end{table*}

\begin{table*}\centering
\scalebox{1}{
\begin{tabular}{l|cccccc|}
\cline{2-7}
                             & \multicolumn{6}{c|}{5min}                                                                                                                                                                                    \\ \cline{2-7} 
                             & \multicolumn{2}{c|}{1 step}                                               & \multicolumn{2}{c|}{5 steps}                                              & \multicolumn{2}{c|}{10 steps}                        \\ \cline{2-7} 
                             & \multicolumn{1}{c|}{F1}             & \multicolumn{1}{c|}{MCC}            & \multicolumn{1}{c|}{F1}             & \multicolumn{1}{c|}{MCC}            & \multicolumn{1}{c|}{F1}             & MCC            \\ \hline
\multicolumn{1}{|l|}{Ori}    & \multicolumn{1}{c|}{0.435}          & \multicolumn{1}{c|}{0.007}          & \multicolumn{1}{c|}{0.452}          & \multicolumn{1}{c|}{0.001}          & \multicolumn{1}{c|}{0.534}          & 0.007          \\ \hline
\multicolumn{1}{|l|}{EMA}    & \multicolumn{1}{c|}{0.494}          & \multicolumn{1}{c|}{0.021}          & \multicolumn{1}{c|}{0.463}          & \multicolumn{1}{c|}{0.004}          & \multicolumn{1}{c|}{0.533}          & 0.021          \\ \hline
\multicolumn{1}{|l|}{DAE}    & \multicolumn{1}{c|}{0.151}          & \multicolumn{1}{c|}{0.029}          & \multicolumn{1}{c|}{0.09}           & \multicolumn{1}{c|}{0.071}          & \multicolumn{1}{c|}{\textbf{0.731}} & 0.029          \\ \hline
\multicolumn{1}{|l|}{VE-SDE} & \multicolumn{1}{c|}{0.668}          & \multicolumn{1}{c|}{0.222}          & \multicolumn{1}{c|}{0.583}          & \multicolumn{1}{c|}{0.152}          & \multicolumn{1}{c|}{0.581}          & 0.082          \\ \hline
\multicolumn{1}{|l|}{VP-SDE} & \multicolumn{1}{c|}{\textbf{0.798}} & \multicolumn{1}{c|}{\textbf{0.313}} & \multicolumn{1}{c|}{\textbf{0.748}} & \multicolumn{1}{c|}{\textbf{0.316}} & \multicolumn{1}{c|}{0.729}          & \textbf{0.217} \\ \hline
\end{tabular}}
\caption{Performance of 5min Dataset on F1-Score and MCC Metrics.}
\label{tab:5min-f1-mcc}
\end{table*}

\begin{figure*}[!htb]  
    \centering  
    \includegraphics[width=\textwidth]{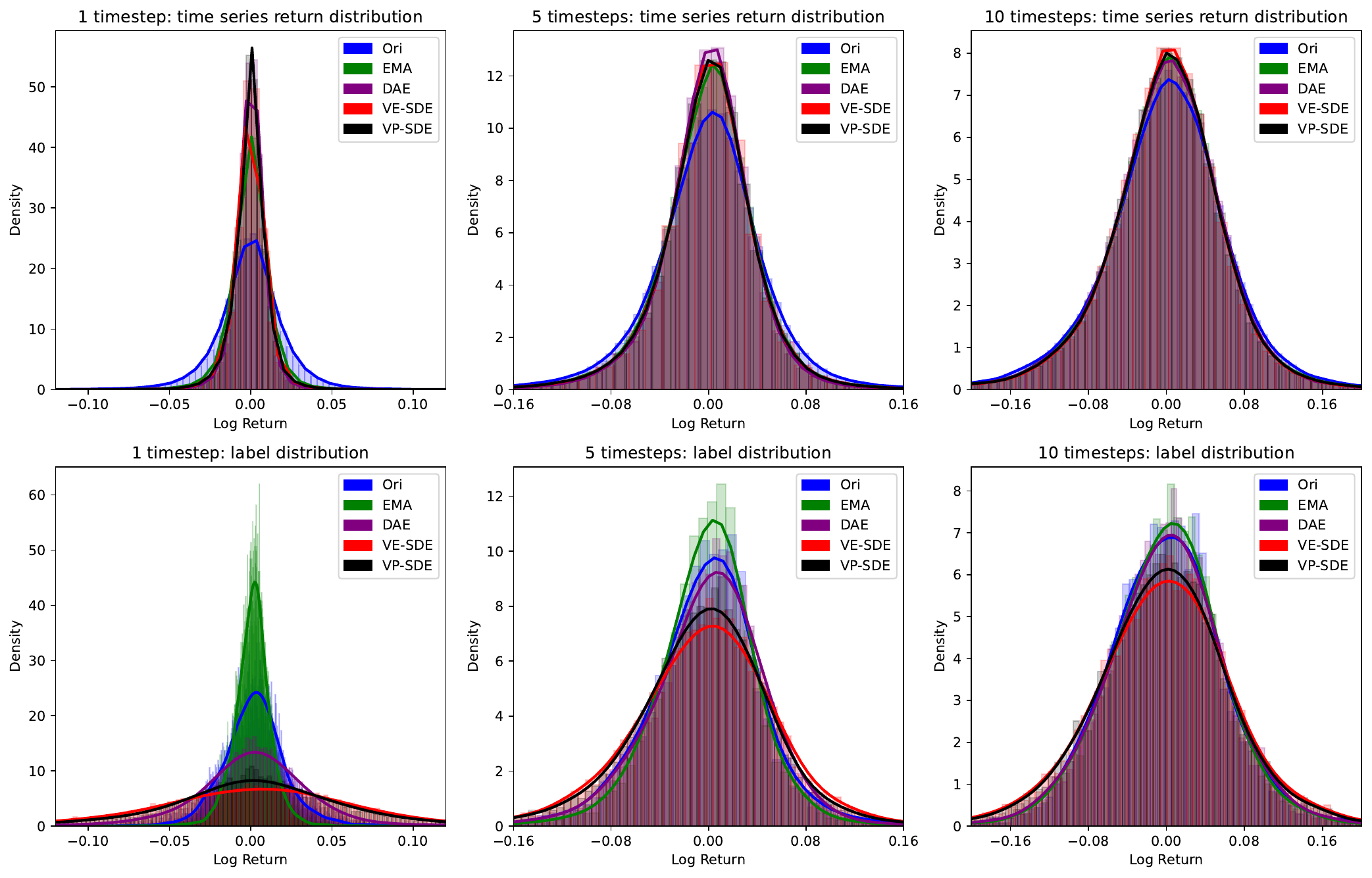}  
    \caption{Time Series Return Distribution and Label Distribution of 1day Dataset.}
    \label{fig:return-distribution}  
\end{figure*}

To further explore why the methods based on the diffusion model perform better than other denoising methods, we take a closer look at the log return distribution of denoised time series and labels. In Figure~\ref{fig:return-distribution}, we draw the return distribution histograms of the 1day dataset, which contains 6 return distribution plots. Considering 1 timestep prediction case, the plot in the top row represents the return distribution obtained by time series $\mathbf{x}_{[t-59, t]}$, where log return is calculated by two sequential value, such as $\log(\frac{\mathbf{x}_t}{\mathbf{x}_{t-1}}), \log(\frac{\mathbf{x}_{t-1}}{\mathbf{x}_{t-2}}), \cdots$; the plot in the bottom row represents the return distribution obtained by two adjacent time series such as $\mathbf{x}_{[t-59, t]}$ and $\mathbf{x}_{[t+1, t+60]}$, where log return is calculated as $\log(\frac{\mathbf{x}_{t+1}}{\mathbf{x}_{t}})$. From Figure~\ref{fig:return-distribution}, we can find that the return distribution of time series becomes smoother while the distribution of labels becomes more scattered after denoising. Therefore, our diffusion model-based denoiser has two functions: ``interior smoothing'' and ``exterior sharpening''. It makes the interior of data less sharp and erases the unimportant details while also making the exterior of data sharper by widening the target distribution. With both functions, the downstream classification task becomes easier.

\subsection{Denoised Time Series For Trading}
\label{subsection:denoised-time-series-for-trading}

To find out if the denoised time series can be used for actual trading, we use the denoised time series to generate trading signals while taking positions and obtaining returns in the original stock time series.


We conduct experiments on 4 stocks during the testing period, which are GOOG, AAPL, MSFT, and AMZN, respectively. We use three metrics to evaluate the trading performance, which are Long-Only Return (LOR), Long-Short Return (LSR) and Number of Trades (NoT). LOR represents the accumulative Long Only Return, where we only take positions when receiving buying signals. LSR represents the accumulative Long-Short Return, where we take positions when receiving buying or selling signals. 
It is better to keep LOR and LSR large but NoT small because frequent trading can lead to high transaction cost and lower return.

\begin{figure*}[!htb]
    \subfigure[MACD Strategy]{
        \centering
        \includegraphics[width=0.5\textwidth]{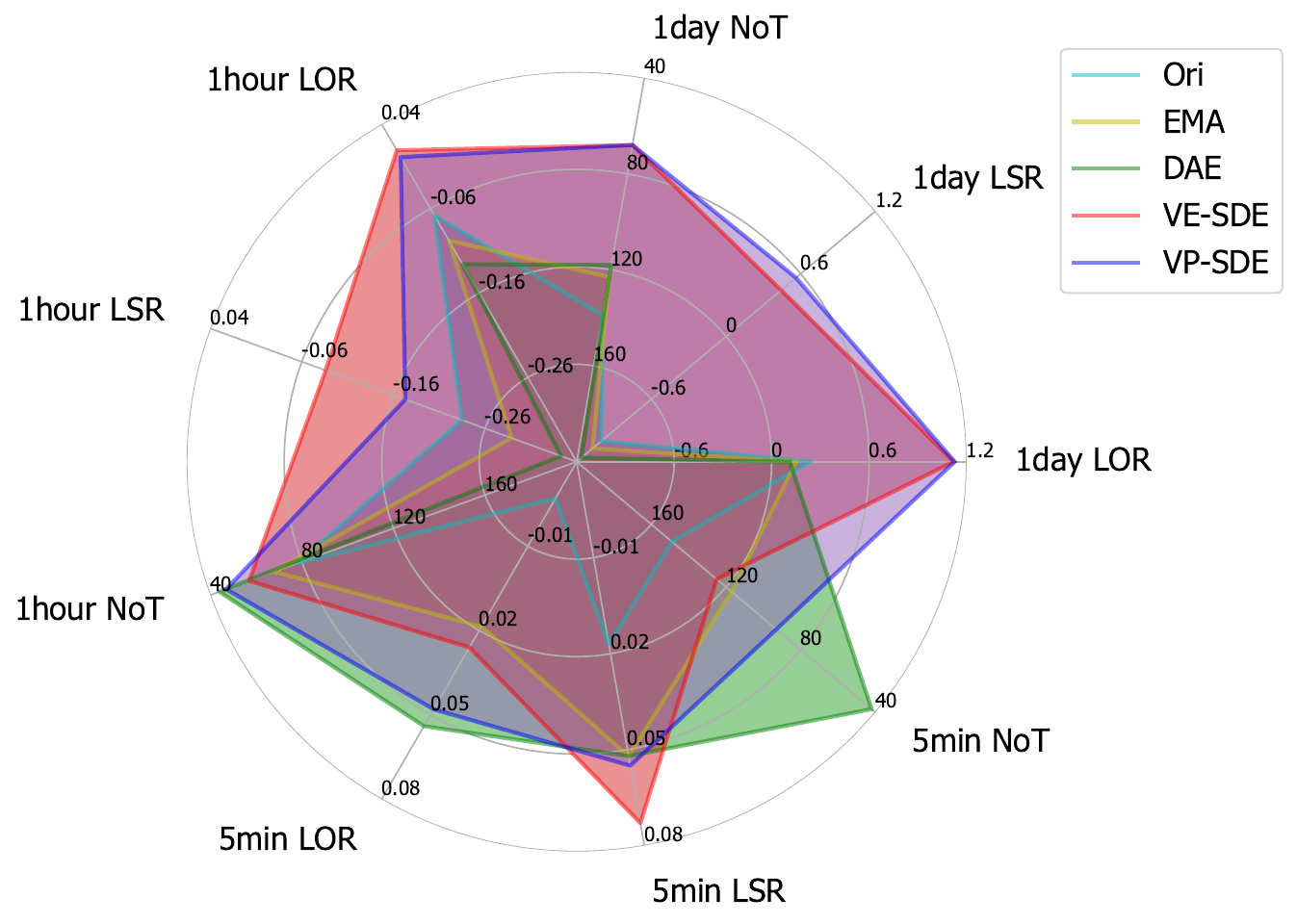}
    }
    \subfigure[Bollinger Strategy]{
        \centering
        \includegraphics[width=0.5\textwidth]{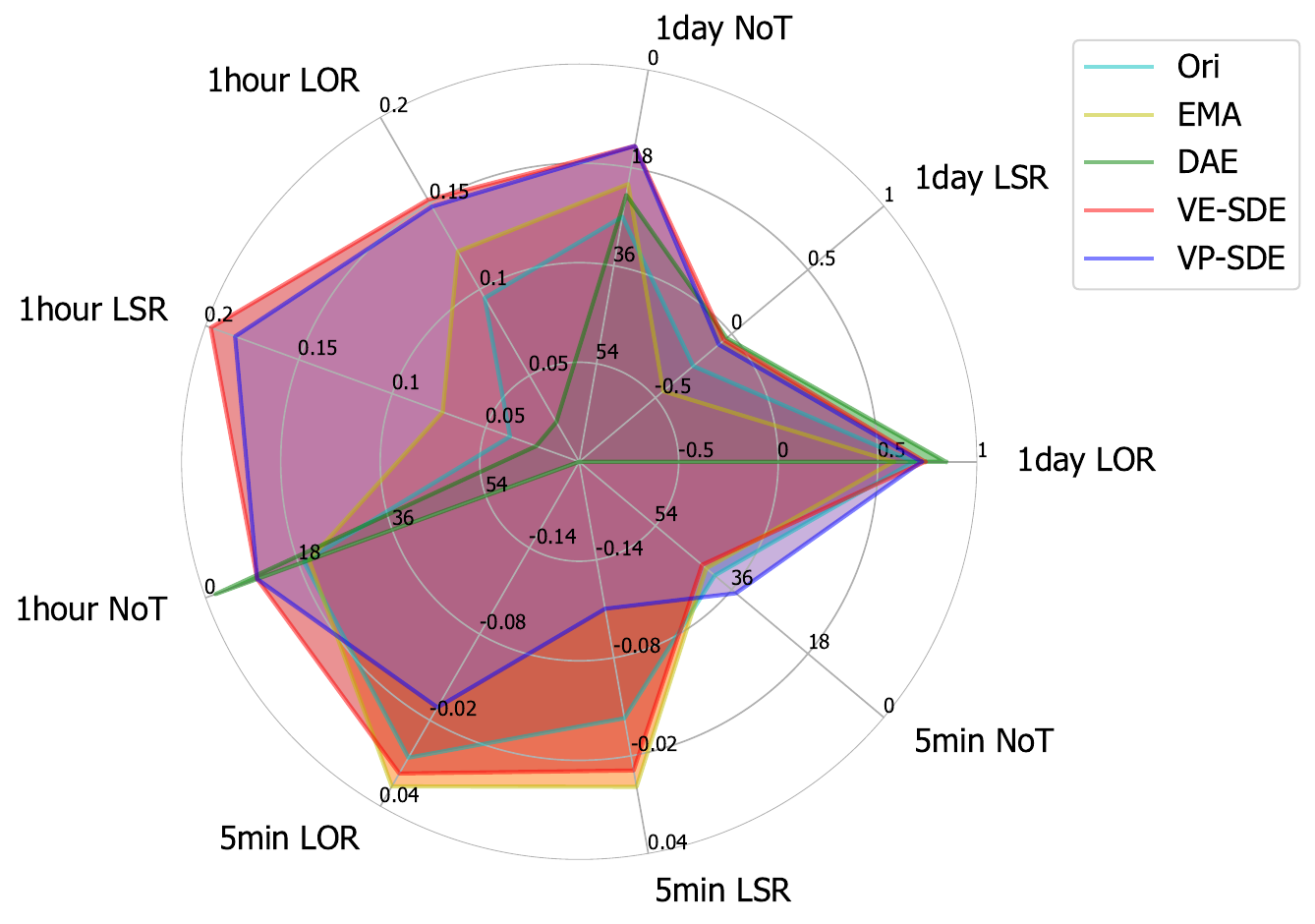}
    }
\caption{Using Signals Generated by Denoised Time Series for Trading.}
\label{fig:time-series-trading}
\end{figure*}

In Figure \ref{fig:time-series-trading}, we show the average trading performance on the 4 stocks by using the standard MACD 
and Bollinger strategies. 
For the MACD strategy, we observe that VE-SDE and VP-SDE perform best on the daily and hourly datasets, while DAE performs slightly better on the 5min dataset. For Bollinger strategy, our diffusion model-based denoising methods beat other models in most cases. We notice that DAE barely provides effective trading signal in 1hour and 5min datasets because the DAE-based denoised time series is too smooth to hit the trading trigger, which can also be supported by the number of DC events in Table \ref{tab:dc-events}.

\subsection{Using the Classifier's Prediction for Trading}

We have demonstrated in Section \ref{subsection:downstream-classification-task} that by using the diffusion model-based denoising method, we can obtain greater performance on the downstream return prediction task. This means that we can predict the denoised market trend when the market is still noisy. In other words, we can tell if the market is noisy or not by comparing the real market trend and our predicted market trend. In this section, we explore how to leverage the information from denoised market trends for trading.

We still use GOOG, AAPL, MSFT, and AMZN for experiments for this. Here, we take 1 timestep prediction as an example to explain our experiments. By using the classifier trained in Section \ref{subsection:downstream-classification-task}, we can take time series $\mathbf{x}_{[t-59, t]}$ as input and obtain a prediction about the market trend on time period $[t, t+1]$. When the input is a denoised time series, the output should be the denoised market trend on $[t, t+1]$, which is called \textit{prediction stage}. We can also observe the true market trend on $[t, t+1]$. By leveraging the information on the prediction stage, we will take positions on $[t+1, t+2]$, which is called \textit{action stage}. We use 1, 5, and 10 timesteps prediction results just as in Section \ref{subsection:downstream-classification-task}. We adopt two opposite trading strategies, one is a following-trend strategy, and the other is a countering-trend strategy. For the following-trend strategy, if the prediction result on the prediction stage shows a positive (negative, respectively) trend, then we long (short, respectively) in the action stage. For the countering-trend strategy, if the prediction result in the prediction stage shows a negative (positive, respectively) trend, then we long (short, respectively) in the action stage. We use four metrics to evaluate the performance of different predictors, which are Long-Only Return (LOR), Long-Only Hit Ratio (LOHR), Long-Short Return (LSR) and Long-Short Hit Ratio (LSHR). Hit ratio, also known as the winning rate, measures the proportion of profitable trades to the total number of trades executed. 

\begin{figure*}[!htb]
    \subfigure[1day Countering-trend Strategy]{
        \centering
        \includegraphics[width=0.5\textwidth]{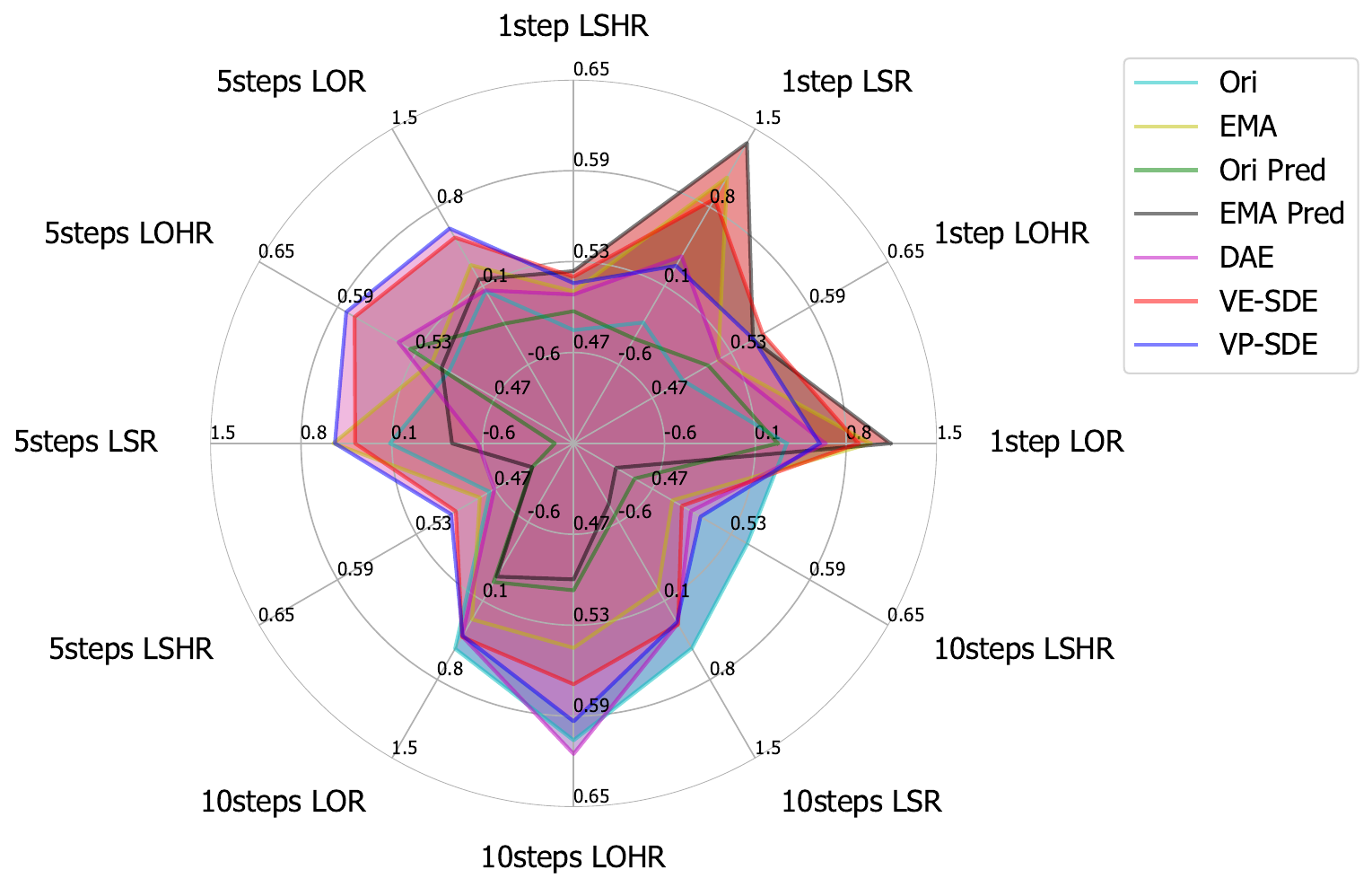}
    }
    \hfill
    \subfigure[1hour Following-trend Strategy]{
        \centering
        \includegraphics[width=0.5\textwidth]{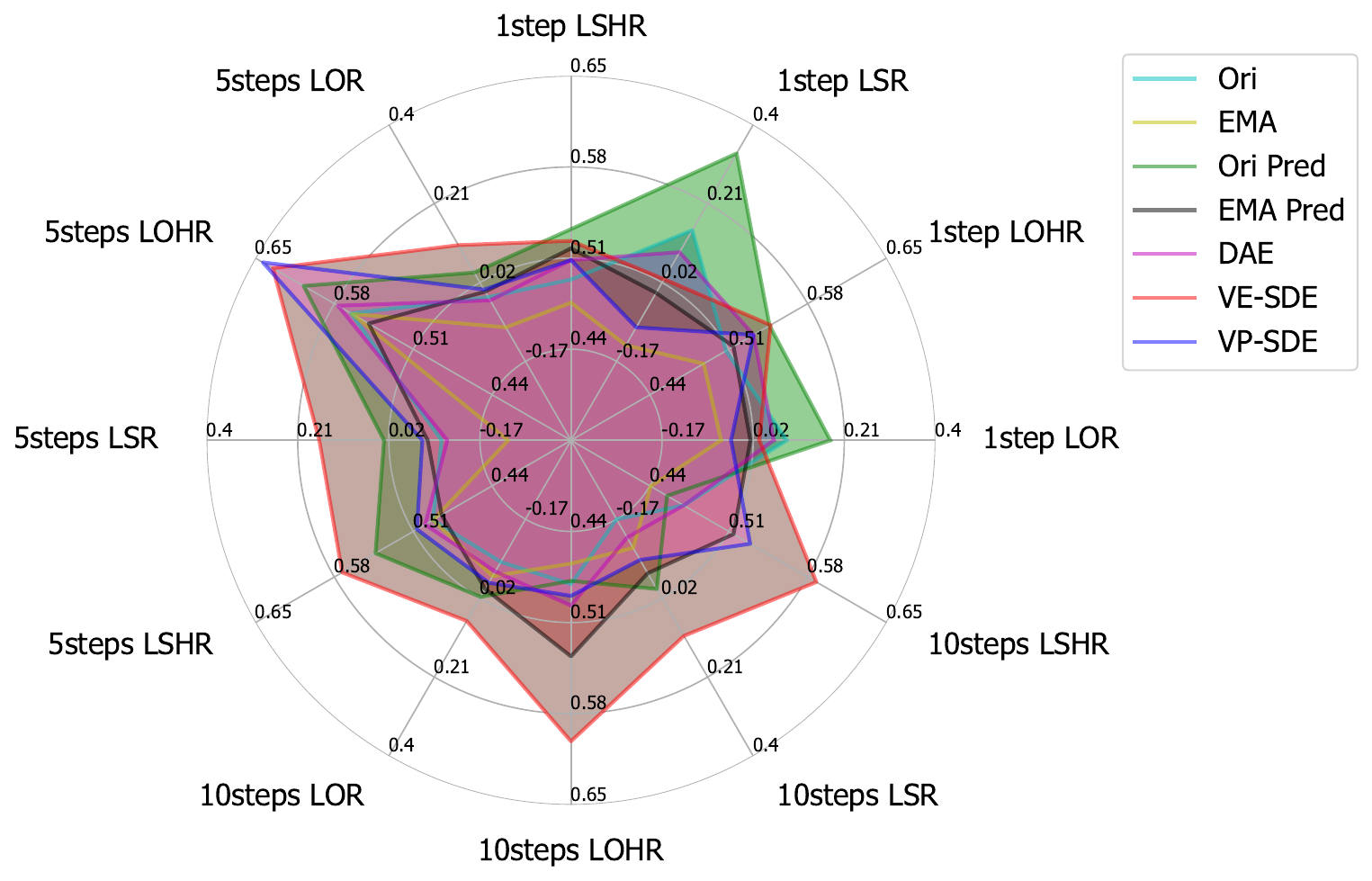}
    }
    
    \hspace{0.25\textwidth}
    \subfigure[5min Countering-trend Strategy]{
        \centering
        \includegraphics[width=0.5\textwidth]{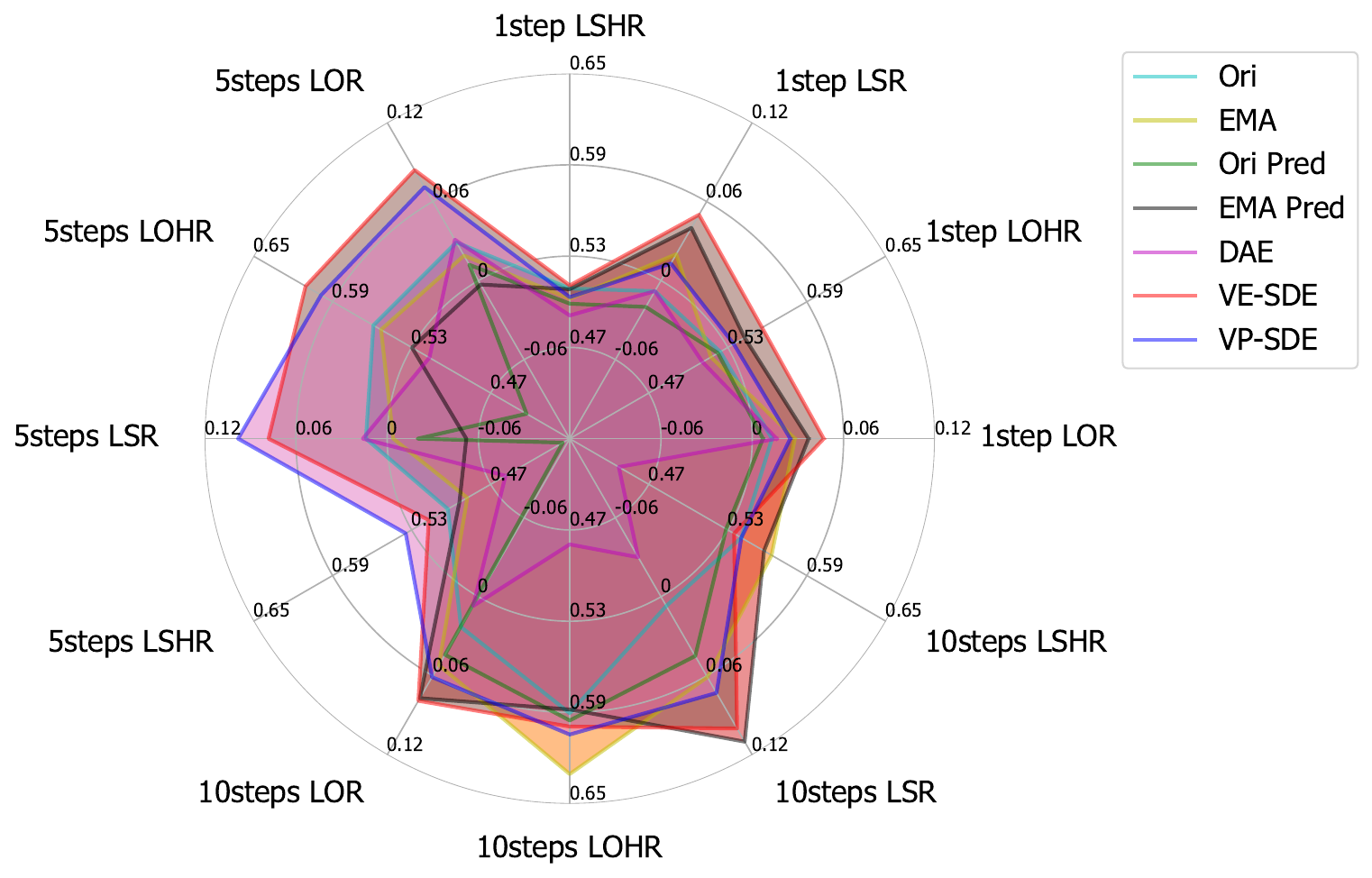}
    }
\caption{Using the 
Prediction Results for Trading.}
\label{fig:classifier-trading}
\end{figure*}

The experiment results are shown in Figure \ref{fig:classifier-trading}. Ori and EMA represent the real market trend observed in the prediction stage, while Ori Pred, EMA Pred, DAE, VE-SDE, and VP-SDE represent the predicted market trend obtained by classifiers. We have two observations from the radar plots. $\mathbf{(1)}$ The denoising methods based on the diffusion model perform better by using countering-trend strategy in the 1day and 5min datasets and by using following-trend strategy in the 1hour dataset. $\mathbf{(2)}$ The denoising methods based on diffusion model perform comparatively better on 5 and 10 timesteps trading than 1 timestep trading. 

We explain the observations from the perspective of how much noise remains in the prediction stage. If the real market contains noise in the prediction stage, the price trend in the prediction stage is also noisy and inaccurate. Then, the market will try to recover to the ``unnoisy'' state in the action stage by countering the current price trend. So, we should take a countering-trend position in the action stage. On the contrary, if the price trend in the prediction stage is not noisy, then the market will likely continue that trend in the action stage, which means we should take the position suggested by following-trend strategy. We believe that the 5min and 1day dataset have more noise than the 1hour dataset. For the 5min dataset, the noise is not smoothed out by the market because the prediction stage is too short. As for the 1day dataset, the prediction stage is so long that new noise may be introduced into the price trend. But for the 1hour dataset, the length of the prediction stage is just enough to balance the market without new noise coming in. The results of DC events can also support this. In Table \ref{tab:dc-events}, we find that VE/VP-SDE has relatively more DC events in the 1hour dataset, where the DC events are preserved well after denoising because the 1hour dataset does not contain much noise in the first place. On the other hand, due to Total Variation and Fourier guidance during the inference process, the time series will carry more information about the long trend. Therefore, VE/VP-SDE performs better on longer timesteps trading tasks.

\section{Conclusion}

This study highlights the potential of using the diffusion model as a financial time series denoiser. The denoised time series exhibits better predictability in downstream return classification tasks and generates more profitable trading signals. By effectively distinguishing between noisy and clear market trends using classifiers trained on denoised time series, the diffusion model provides more accurate market predictions and improved trading decisions. The findings suggest that diffusion models can significantly enhance the signal-to-noise ratio of financial time series, thereby improving the reliability of predictive models and trading strategies.

\bibliographystyle{unsrtnat}  
\bibliography{references}

\end{document}